\documentclass[a4paper,11pt]{article}

\usepackage{fourier}
\usepackage{color}
\usepackage{graphicx}
\usepackage{url}
\usepackage[affil-it]{authblk}
\usepackage{amsmath}
\usepackage{wrapfig}
\usepackage{subcaption}
\usepackage{gensymb}

\usepackage[T1]{fontenc}
\usepackage{times}

\begin{document}

\title{\textbf{YUDO}: \textbf{Y}OLO for \textbf{U}niform \textbf{D}irected \textbf{O}bject Detection}

\author{Đorđe Nedeljković}
\affil{Independent Researcher}
\date{}
\maketitle
\thispagestyle{empty}

\begin{abstract}
This paper presents an efficient way of detecting directed objects by predicting their center coordinates
and direction angle. Since the objects are of uniform size, the proposed model works without predicting the object's width and height. The dataset used for this problem is presented in
Honeybee Segmentation and Tracking Datasets project \cite{Bozek2017TowardsDO}. One of the contributions
of this work is an examination of the ability of the standard real-time object detection architecture like YoloV7 \cite{Wang2022YOLOv7TB} to be customized
for position and direction detection. A very efficient, tiny version of the architecture is used in this approach. Moreover, only one of three detection heads without anchors is sufficient for this task.
We also introduce the extended Skew Intersection over Union (SkewIoU) \cite{Ma2017ArbitraryOrientedST} calculation for rotated boxes - directed IoU (\textbf{DirIoU}), which includes an absolute angle difference.
\textbf{DirIoU} is used both in the matching procedure of target and predicted bounding boxes for mAP calculation, and in the NMS filtering procedure.
The code and models are available at \url{https://github.com/djordjened92/yudo}.

\end{abstract}

\section{Introduction}
\begin{wrapfigure}[23]{r}{0.6\textwidth}
  \vspace{-25pt}
  \begin{center}
    \includegraphics[width=0.6\textwidth, height=0.6\textwidth]{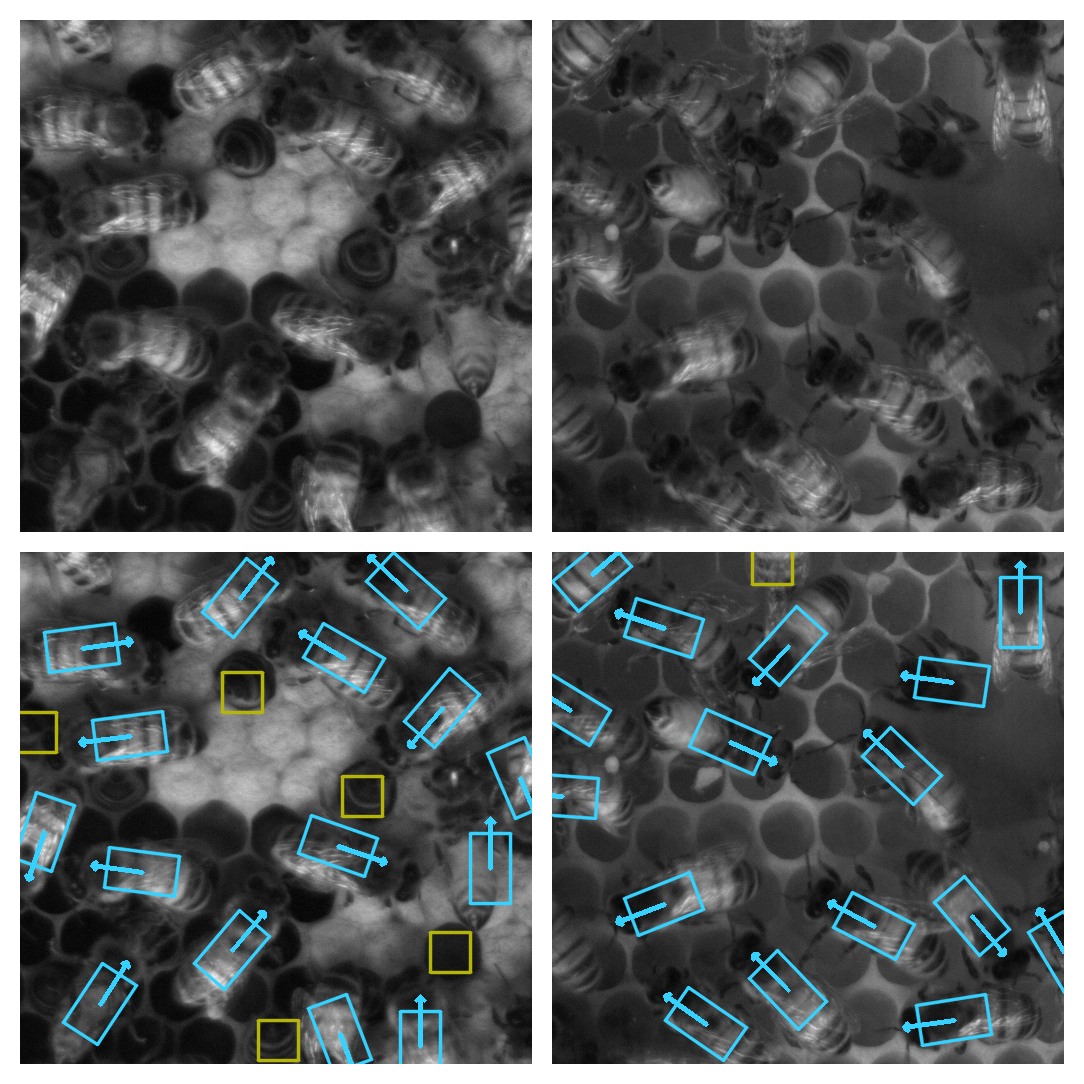}
  \end{center}
\vspace{-20pt}
  \caption{Dataset image samples with annotation visualizations.}
  \vspace{-0pt}
\end{wrapfigure}

The original dataset from the paper \cite{Bozek2017TowardsDO}, which inspired this work was
intended to be used for resolving the general multi-object tracking in a crowded environment like a beehive.
Each bee was manually annotated with $(x, y, t, \theta)$, where $x$ and $y$ represent coordinates of the body center, $t$ class label
(1 when the full body is visible and 2 when the bee is inside a comb cell) and $\theta$ is the angle of body direction against the vertical
bottom-up axis with values from the range $[0, 2\pi)$ calculated clockwise. An angle is annotated only for the bee class ($t = 1$) and ignored
for the abdomen class ($t = 2$). In the original work, the localization of the tracked honeybee bodies was implemented through a segmentation approach using U-Net \cite{ronneberger2015unet}, in combination
with angle distance loss and temporal component, so authors generated an approximation of bee body size annotations (around the annotated center) using
ellipses for $t = 1$ with semi-minor axis $r1 = 20$ pixels and semi-major axis $r2 = 35$ pixels, and for $t = 2$ the model is a circle with $r = 20$.
We aim to show that modern object detection architectures can predict the center and direction angle of uniformly sized objects in a dense configuration.
Therefore, the neural network is not provided with object size during the training time. The beforementioned approximation of the bee body is transformed into a more suitable
bounding box format which encloses an ellipse or circle for the abdomen label. These bounding boxes are used in the process of non-maximum suppression filtering of predictions
and prediction-target matching in evaluation time. Two samples of images (top) and their annotations (bottom) are shown in Figure 1. The blue arrowed boxes
represent the bee class and the yellow squares represent the abdomen class.

\section{Related Work}
Object detection is one of the most important and one of the most comperehensively developed research areas in computer vision, and nowadays
deep learning methods are a dominant approach for this task. According to the alignment of the labeled and detected boxes, this field of research
is divided into more general horizontal detection and rotated object detection. Horizontal object detection, by which all boxes are aligned to horizontal
and vertical axes, is more suitable for natural scene images. For the purpose of more accurate object positioning, like scene text detection or aerial image analysis, rotated detection is a more suitable approach.

\subsection{Horizontal and Rotated Object Detection}
\paragraph{Horizontal Object Detection:}
Horizontal object detection is the default object detection subfield which assumes horizontal bounding boxes as object representations.
The dominant backbones so far in this field are ConvNets \cite{6795724} of various forms.\\\\
\textit{Two stage detection} is a family of approaches which are dominantly region based. In the first stage, they generate category-independent proposals of bounding boxes, and then after feature extraction of those regions, they apply classification
and regression in the second stage. Fast RCNN \cite{Girshick2015FastR}, Faster RCNN \cite{Ren2015FasterRT} and R-FCN \cite{Dai2016RFCNOD} belong to this category and they
don't exploit the hierarchical structure of ConvNets but rely on the single feature map instead, in both stages. FPN \cite{Lin2016FeaturePN}\\ is also a two stage approach, but
it exploits all stages of the hierarchical backbone with top-down and lateral connections. The FPN design inspired many recent object detection architectures.\\\\
\textit{Single stage detection} methods aim to find regions of interest, regress coordinates, and apply classification in one neural network pass. Due to their efficiency, most of
the real-time solutions rely on this approach. SSD \cite{Liu2015SSDSS} generate class scores and bounding box parameters per feature map location on different scales of convolution layers.
Therefore, this has been one of the first methods leveraging the hierarchical nature of ConvNets. RetinaNet \cite{Lin2017FocalLF} is trying to address background/foreground class imbalance by implementing the FocalLoss.
The most popular family of single stage detection is YOLO \cite{Redmon2015YouOL}. Although there has been a lot of progress in the YOLO-type detectors over time, the principle of coordinates regression and class scores
prediction directly from feature map(s) of ConvNets remains throughout all versions.\\\\
Most of the mentioned approaches use bounding box suggestions called anchors in the prediction stage, but there are also so-called anchor-free approaches like FCOS \cite{Tian2019FCOSFC}, CornerNet \cite{Law2018CornerNetDO}, CenterNet \cite{Zhou2019ObjectsAP}.
Nowadays, transformer is the most popular and widely spread architecture. Object detection is no exception, so DETR \cite{Carion2020EndtoEndOD} and ViTDet \cite{Li2022ExploringPV} are representatives of this category.

\paragraph{YoloV7:}
Besides a few trainable bag-of-freebies, which are tricks used in training proposed by other authors, this work introduced some original architectural improvements. The YoloV7 \cite{Wang2022YOLOv7TB} architecture belongs to the concatenation-based type of model,
which usually consists of computational and transition blocks. The computational blocks are the main components which are aimed to distillate qualitative feature maps, using parallel branches of inference. The results are aggregated using a concatenation.
The transition layers are helper elements, usually built from convolution and pooling operations, used to maintain output resolution and the number of channels. Inspired by VoVNet \cite{Lee2019AnEA} architecture, the authors suggested an improved computational block called E-ELAN (Extended Efficient Layer Aggregation Networks).
A group convolution is used to expand the channel and cardinality of computational blocks. Output feature maps of these groups are shuffled, concatenated, and finally added.\\
An additional contribution of this approach is the specific scaling of models. Increasing depth in a concatenation-based model implies that the output width of a computational block also increases. YoloV7 \cite{Wang2022YOLOv7TB} deals with this by scaling up the width of transition layers.\\
There are few designed basic models of YoloV7 (for edge GPU, normal GPU, and cloud GPU). We tried to apply our approach to the most efficient, edge GPU version of architecture -  YOLOv7-tiny.

\paragraph{Rotated Object Detection:}
The mainstream approaches of rotated detection are mostly adapting the horizontal detection paradigm by representing the bounding boxes with five parameters: coordinates of the bounding box center, bounding box width, bounding box height, and rotation angle
- $(x, y, w, h, \theta)$. The rotation angle can be determined by the $x$-axis and closest rectangle edge in the $90^{\degree}$ range $\theta \in [-90^{\degree},0^{\degree})$, or by the long side of the rectangle and the $x$-axis in the  $180^{\degree}$ range $\theta \in [-90^{\degree},90^{\degree})$.
The notation is that the clockwise direction is negative.\\
The region based approaches usually regress these five parameters, implementing different variants of $l_n$-norm losses (such as smooth $l_1$ loss) like
R-DFPN \cite{Yang2018AutomaticSD}, R\textsuperscript{3}Det \cite{Yang2019R3DetRS}, RSDet \cite{Qian2019LearningML}, CSL \cite{Yang2020OnTA}, or exploiting differentiable approximation of IoU loss like SCRDet \cite{yang2019scrdet}, GWD (Gaussian Wasserstein distance) \cite{Yang2021RethinkingRO}.\\
Either directly or implicitly, all these methods rely on the regression of the rotation bounding:

\begin{equation} \label{eq:rot_box_offset}
\begin{split}
  t_x &= (x - x_a)/w_a, t_y = (y - y_a)/h_a \\
  t_w &= log(w/w_a), t_h = log(h/h_a), t_\theta = f(\theta - \theta_a) \\
  t^{'}_x &= (x^{'} - x_a)/w_a, t^{'}_y = (y^{'} - y_a)/h_a \\
  t^{'}_w &= log(w^{'}/w_a), t^{'}_h = log(h^{'}/h_a), t^{'}_\theta = f(\theta^{'} - \theta_a)
\end{split}
\end{equation}
where $x$, $x_a$, $x^{'}$ are for the ground-truth box, anchor box and predicted box, respectively (likewise $y$, $w$, $h$, $\theta$). The function
$f(\cdot)$ is usually used to deal with angular periodicity such as trigonometric functions, modulo (like for the $l^{5p}_{mr}$ loss in the RSDet \cite{Qian2019LearningML}), etc.
The regression loss usually has a general form:
\begin{equation} \label{eq:rot_box_reg}
  L_{reg} = l_n-norm(\Delta{t_x}, \Delta{t_y}, \Delta{t_w}, \Delta{t_h}, \Delta{t_\theta})
\end{equation}
where $\Delta{t_j}=|t_j - t^{'}_j|$ for $j \in \{x, y, w, h, \theta\}$.

\begin{wrapfigure}[11]{r}{0.58\textwidth}
  \vspace{-50pt}
  \begin{center}
    \includegraphics[width=0.58\textwidth, height=0.34\textwidth]{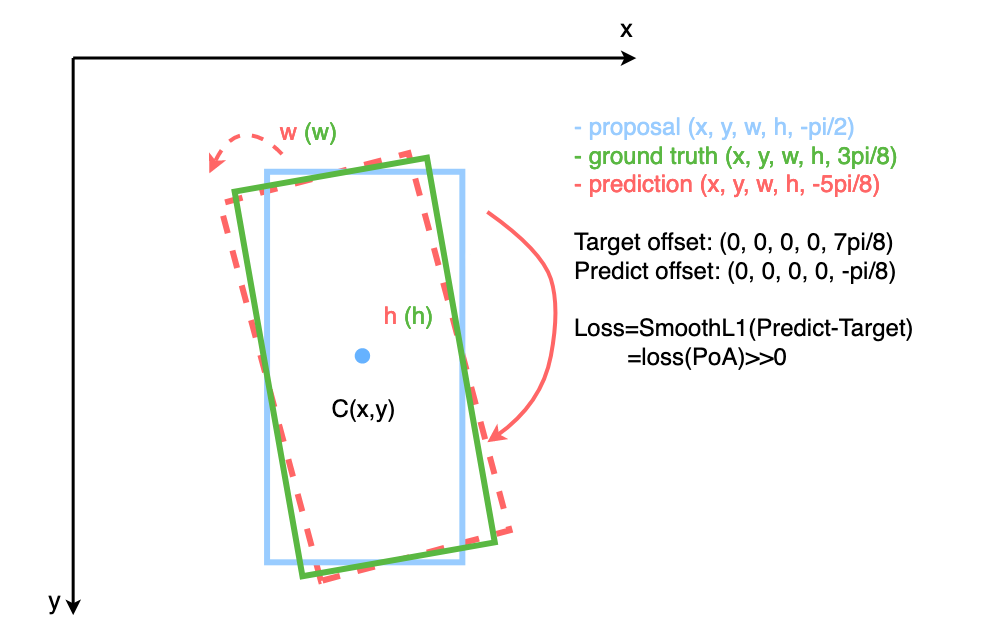}
  \end{center}
  \vspace{-20pt}
  \caption{The boundary discontinuity caused by PoA on the $180^{\degree}$ range example \cite{Yang2020OnTA}.}
  \vspace{-0pt}
\end{wrapfigure}

All aforementioned methods contain this loss component, extended in different ways
to address \textbf{boundary discontinuity problems} mainly caused by the periodicity of angular (PoA). Put simply, when the target and predicted angle are close to the
opposite ends of the range, the regression loss suddenly increases. In the case of $180^{\degree}$ range shown in Figure 2, the ground-truth bounding box is close to the
$\pi/2$, and the predicted box is close to the $-\pi/2$. As rotated object detection is agnostic to the direction, then these 2 boxes are visually very similar despite of high 
Smooth L1 loss.

\section{Directed Object Detection}
Since the standard rotated object detection is agnostic to the direction of objects, we introduce a method for directed object detection in this work. The angle of a directed object is measured relative to only one certain edge, in contrast to rotated object detection.
For the main backbone architecture, we use YOLOv7-tiny \cite{Wang2022YOLOv7TB}.

\subsection{Bounding Box Parametrization}
The bounding box model of a honeybee is shown in Figure 3. The angle of the bee's direction $\theta$ is determined by the vertical axis (bottom-up orientation) and the bee's head position.
Its value can be from the range $[0, 2\pi)$ calculated in the clockwise direction. Central point coordinates are $(x,y)$.

\subsection{Regression method}
The concept of YOLO architecture is to map detection head output into the grid which covers the input image. Each cell provides $N_a$ prediction vectors where each of them represents one predicted bounding box, and $N_a$ is the number of anchors.
Thus each cell's output nodes related to the coordinates $(x,y)$ represent the offset from the cell top-left corner. These values are in the normalized form, relative to the width and height of the cell's receptive field.
It's intuitive why these nodes have a sigmoidal activation function since the possible position of the predicted bounding box
is from the top-left to the bottom-right border point of the cell. Our simplifications:
\begin{wrapfigure}{r}{0.42\textwidth}
  \vspace{-20pt}
  \begin{center}
    \includegraphics[width=0.42\textwidth, height=0.4\textwidth]{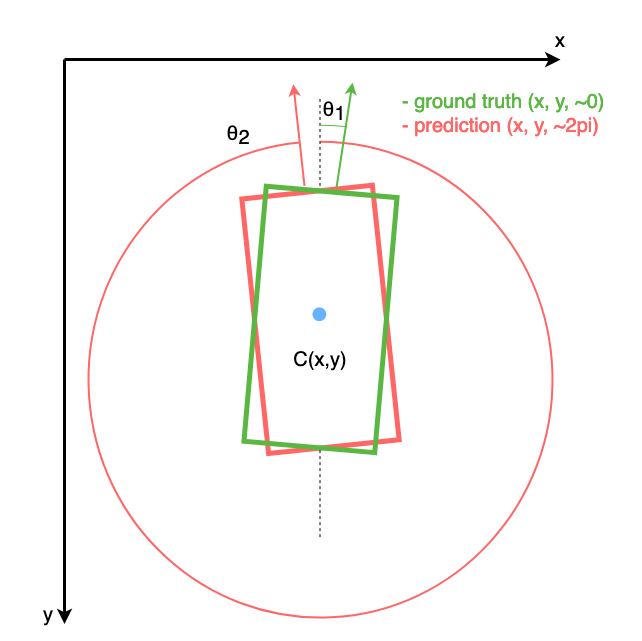}
  \end{center}
  \vspace{-10pt}
  \caption{Bounding box model of honeybee.}
\end{wrapfigure}
\begin{itemize}
  \item[$-$] $(w,h)$ nodes for bounding box width and height are omitted from the prediction vector since those values are predefined and uniform.
  \item[$-$] We used an anchor-free approach where each cell has one possible prediction, which is justified when overlapping of objects is not expected and if the output grid is dense enough in order not to miss prediction in the crowded parts of the scene.
  \item[$-$] Instead of three detection heads (with different grid sizes and assigned sets of anchors) only one is kept with grid size 16x16.
\end{itemize}
Before loss calculation, all ground-truth bounding boxes are assigned to the specific cell according to the central point position. The regression loss for the central point is the mean squared error:

\begin{equation} \label{eq:reg_xy_loss}
  L_{xy} = \frac{1}{N}\sum_{i=1}^{N}((\hat{x}_i-x_i)^2 + (\hat{y}_i-y_i)^2)
\end{equation}
where $t_i$ is the relative coordinate offset of ground-truth box, $\hat{t}_i$ is the coordinate of assigned prediction for $t \in \{x,y\}$.
$N$ is total number of bounding boxes.\\

Alongside $(x,y)$ nodes in the output vector for each cell, there is a dedicated node for the angle $\theta$ prediction. The aforementioned \textbf{boundary discontinuity problem} caused by PoA appears in this setup, too.
An example is presented in Figure 3, where the target bounding box has the angle $\theta_1\approx0$ and the predicted angle is $\theta_2\approx2\pi$. In the standard $L_1$ loss, the absolute difference $\approx2pi$ would cause sudden raise of regression loss,
although it is obvious these bounding boxes have close angles of direction. One possible way to deal with this issue is to use cosine distance: $1 - \cos(\Delta\theta)$, which has the lowest value of 0 for the case of a small difference in direction angles,
and a maximum value of 2 in the case when directions are opposite. The output angle node has ReLU activation function and afterward update of value $\theta=\theta(mod\:2\pi)$. The cosine distance loss has the form:

\begin{equation} \label{eq:reg_theta_loss}
  L_{\theta} = \frac{1}{N}\sum_{i=1}^{N}(1 - \cos(\hat{\theta}_i-\theta_i))
\end{equation}
where $\theta$ is the target angle of direction, $\hat{\theta}$ is the predicted angle and $N$ is number of boxes.\\\\
Therefore, the output vector for each grid cell is defined as $(x,y,\theta,obj,abd_{cls},bee_{cls})$. The node $obj$ represents the objectness node responsible for the sureness of prediction existence, and nodes $abd_{cls},bee_{cls}$ are classification nodes
for the abdomen and regular bee class respectively. Activations for classes are independent logistic classifiers. The total loss is defined as:

\begin{equation} \label{eq:total_loss}
  L = \lambda_{xy}L_{xy} + \lambda_{\theta}L_{\theta} + \lambda_{cls}L_{cls} + \lambda_{obj}L_{obj}
\end{equation}
where $\lambda_{i}$ is a weight for specific loss in the total sum. $L_{cls}$ and $L_{obj}$ are classification and objectness losses in the form of standard binary cross-entropy losses.
The loss and activations related to objectness and classification are kept from the original implementation of YoloV7 \cite{Wang2022YOLOv7TB}.

\subsection{Directed IoU}
\begin{wrapfigure}[13]{r}{0.5\textwidth}
  \vspace{-30pt}
  \begin{center}
    \includegraphics[width=0.5\textwidth, height=0.33\textwidth]{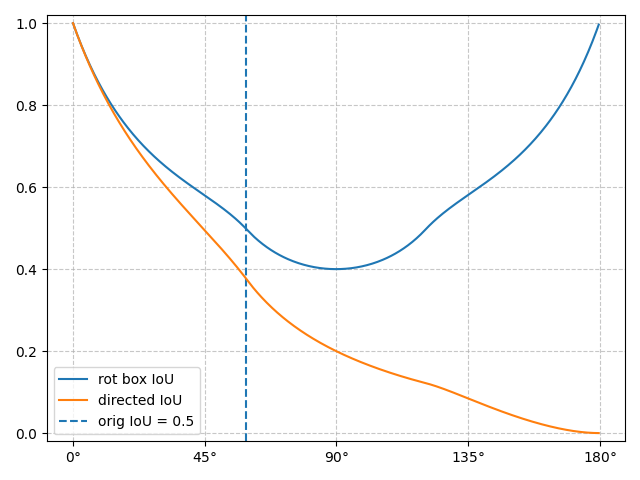}
  \end{center}
  \vspace{-20pt}
  \caption{Rotated bounding box IoU and Directed IoU.}
\end{wrapfigure}
Skew Intersection over Union (SkewIoU), introduced by \cite{Ma2017ArbitraryOrientedST}, is an IoU version adapted for rotated bounding boxes. The premise of this property adopted from horizontal bounding boxes is the same,
with suggestions for intersection area calculation.
It includes the difference in bounding box orientation, thus it is sufficient for describing oriented bounding box detection.
However, this formulation of IoU suffers from a lack of explicit taking into account the difference in \textbf{direction angle} of a bounding box which has a certain "head" edge, for which the direction angle is calculated.
We upgraded the IoU for rotated bounding boxes in Equation (\ref{eq:dir_iou})
with the correction factor, which emphasizes the difference in direction.
\begin{equation} \label{eq:dir_iou}
\begin{split}
  &DirCorr(\Delta\theta) = \frac{1 + cos(\Delta\theta)}{2}\\
  &DirIoU = IoU\cdot DirCorr
\end{split}
\end{equation}
For the calculation of the rotated bounding box IoU, we used detectronv2 \cite{wu2019detectron2} implementation. Figure 4 depicts the effect of the correction factor ($DirCorr(\theta)$ in Equation \ref{eq:dir_iou}) on the rotated bounding box IoU.
Two bounding boxes (in this case two bee class boxes with $w=40, h=70$), which have aligned center positions, reach maximal $IoU=1$ for an angle difference of $0^{\degree}$ or $180^{\degree}$. To emphasize the importance of direction,
our $DirIoU$ has maximal value only when boxes are aligned and the angle difference is $0^{\degree}$. The lowest value of $DirIoU$ is reached when boxes are not intersecting or when their directions are opposite. In Figure 4 $DirIoU$
has a value 0 for the angle difference of $180^{\degree}$, where boxes are overlapping completely, but their directions are opposite. It is evident that this criterion of IoU is more rigid, so we used the threshold of 0.3 instead of the default value of 0.5 for mAP calculation and NMS filtering.

\section{Experiment}
\paragraph{Dataset:}
As suggested in the original work \cite{Bozek2017TowardsDO}, we randomly sampled test data in equal proportions from both image pools of 30FPS and 70FPS recordings.
The model is trained on 13908 images of size 512x512 and validated on 1392 images of the same size. We applied a few augmentation techniques during the training, such as vertical and horizontal flips, 2x2 mosaic image arrangement,
and HSV random parameter changes. Since the original images have blurred regions, we applied a sharpening filter also. On image loading, we filter it by 3x3 kernel with center value 9 and -1 for other cells.
\paragraph{Model and Training:}
We customized the YoloV7 \cite{Wang2022YOLOv7TB} tiny architecture as a detection model. As mentioned before, this customization implies an anchor-free version of the model where only one of three detection heads is kept.
This neural network has \textasciitilde 6 million parameters or 13.1 GFLOPS. The model is trained for 200 epochs with batch size 4, applying the one-cycle learning rate schedule \cite{He2018BagOT} with an initial learning rate of 0.001. 
Initial model weights are from the pre-trained model available at the official github repository. We use Stochastic Gradient Descent as an optimizer and Exponential Moving Average (EMA) of model weights update for stabilized training.
Chosen loss weights are: $\lambda_{xy}=0.1,\lambda_{\theta}=0.1,\lambda_{cls}=0.3,\lambda_{obj}=1.0$. The training is conducted using the GeForce RTX2070 Super GPU.
\paragraph{Results:}
The best checkpoint mAP@30 result is \textbf{70.5}, and class-specific metrics are presented in Table \ref{tab:metrics}. For the bee class, we're applying directed object detection with the center and the angle prediction, but for the abdomen class, the center is predicted only.
So in terms of the result discussion, the bee class is much more important, where obtained numbers are pretty satisfying. The abdomen class appears problematic in terms of results, and one of the causes might be non-consistent annotations noticed during the error analysis.
\begin{table}[!h]
\begin{center}
\begin{tabular}{|c|c|c|c|c|}
  \hline
   & Labels & Precision & Recall & AP\textsubscript{30}\\
  \hline
  bee & 21163 & 82.3 & 88.2 & 85.1\\
  \hline
  abdomen & 2940 & 58.5 & 60.0 & 55.9\\
  \hline
\end{tabular}
\end{center}
\vspace{-20pt}
\caption{Detection results.} \label{tab:metrics}
  \vspace{-10pt}
\end{table}

\section{Conclusion}
We propose the detection method for directed, uniform objects and the corresponding customization of modern horizontal object detectors. This approach meets the requirements of the problem with complete omission of objects' width and height. The directed bee class is detected with mAP@30=\textbf{85.1} using only one YOLOv7-tiny \cite{Wang2022YOLOv7TB} detection head in simplified anchor-free form.
The proposed angle loss component overcomes the problem of angle periodicity.\\
In addition, we define a proper metric (DirIoU) for overlapping measurement of directed objects. It emphasizes the angle difference, introducing an additional factor alongside the standard IoU, which is focused on the object areas only.
That periodic function cancels completely any area overlapping if the directions are opposite ($\Delta\theta\approx\pi$) and keeps the IoU value in the case when directions are similar ($\Delta\theta\approx0$).

\newpage
\appendix

\section{Examples of detections and problematic abdomen annotations}
\begin{figure}[h]
  \begin{minipage}{.45\textwidth}
    \begin{center}
      \includegraphics[scale=0.15]{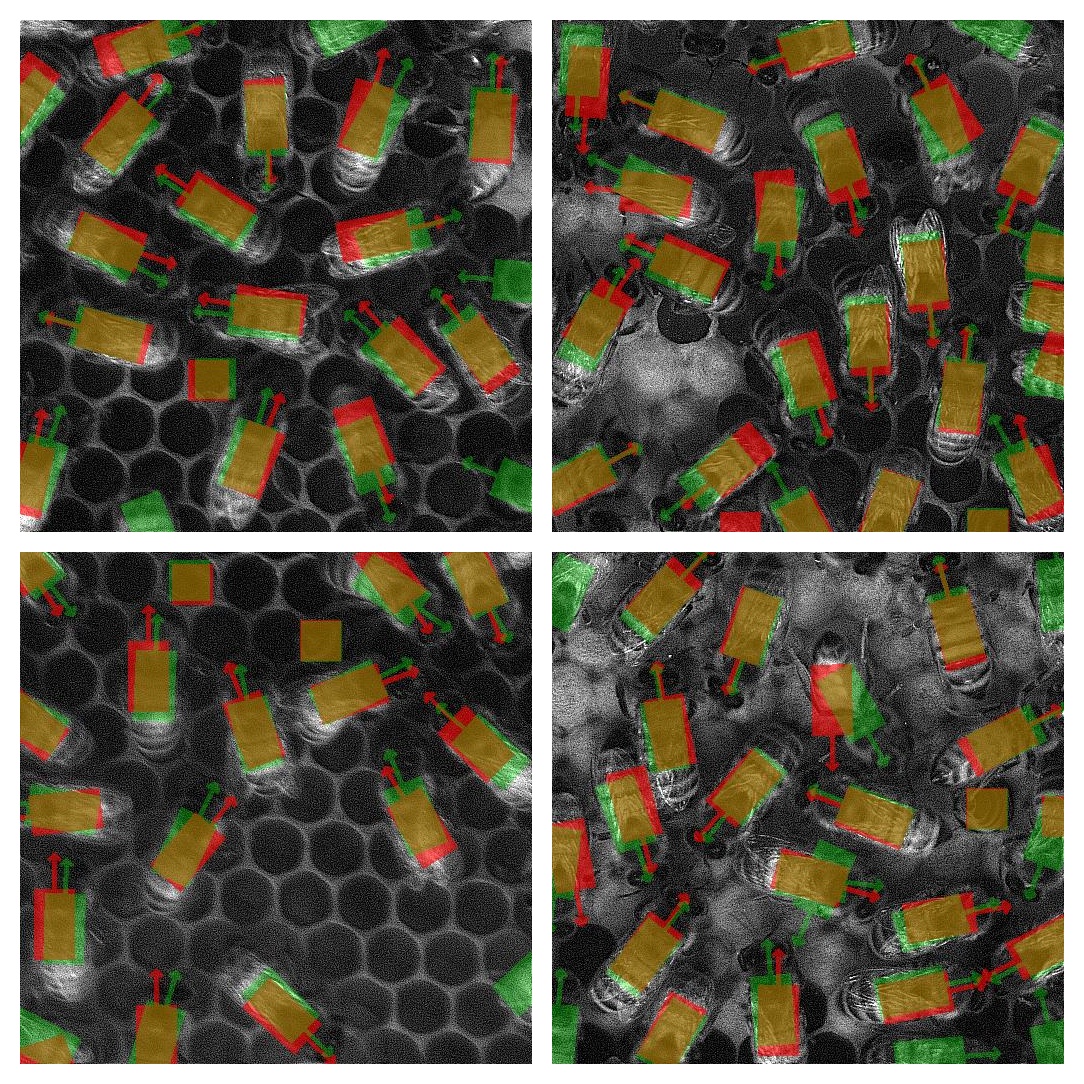}
    \end{center}
    \vspace{-20pt}
    \caption[5]{Detection examples. Arrowed rectangles are bee class and squares are abdomen class. Red color represents target and green is prediction (overlappings are yellow).}
    \vspace{-0pt}
  \end{minipage}\hfil%
  \begin{minipage}{.45\textwidth}
    \begin{center}
      \includegraphics[scale=0.15]{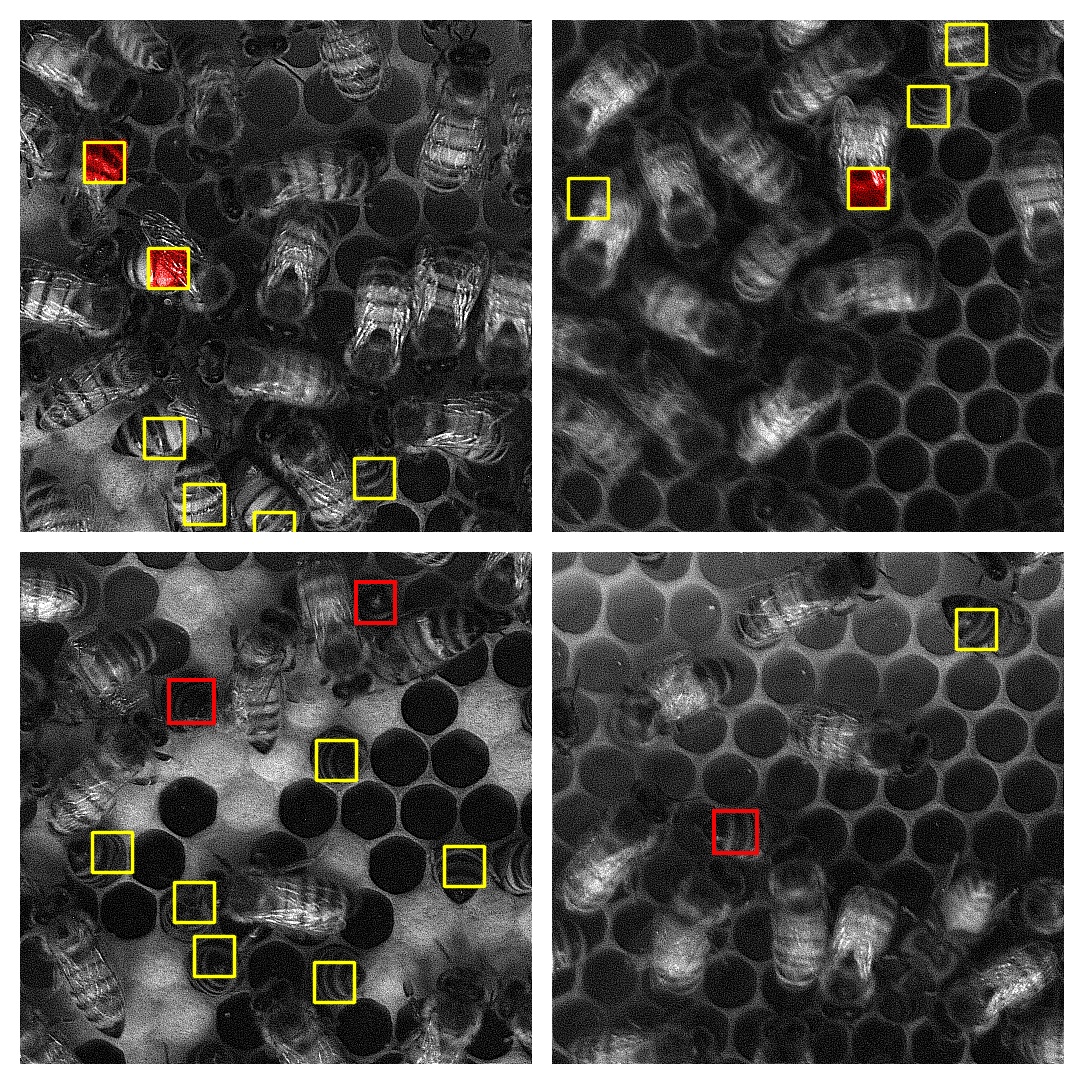}
    \end{center}
    \vspace{-20pt}
    \caption[5]{Problematic abdomen annotations. On the top two images wrong abdomen labels contain red fill. Bottom images have missed abdomen labels shown with red squares.}
    \vspace{-0pt}
  \end{minipage}
\end{figure}

\bibliographystyle{apalike}

\bibliography{imvip}

\end{document}